\documentclass[]{fairmeta}

\title{A Simple Baseline for Streaming Video Understanding}

\author[]{Yujiao Shen}
\author[]{Shulin Tian}
\author[]{Jingkang Yang}
\author[]{Ziwei Liu}
\affiliation[]{S-Lab, Nanyang Technological University}

\date{April 1, 2026}
\metadata[Codebase]{\href{https://github.com/EvolvingLMMs-Lab/SimpleStream}{\texttt{https://github.com/EvolvingLMMs-Lab/SimpleStream}}}
\metadata[Project Page]{\href{https://simple-stream.github.io/}{\texttt{https://simple-stream.github.io/}}}

\abstract{Recent streaming video understanding methods increasingly rely on
complex memory mechanisms to handle long video streams. We challenge
this trend with a simple finding: a sliding-window baseline that feeds
only the most recent $N$ frames to an off-the-shelf VLM already matches
or surpasses published streaming models. We formalize this baseline as
\textsc{SimpleStream} and evaluate it against 13 major offline and
online video LLM baselines on OVO-Bench and StreamingBench. Despite its
simplicity, \textsc{SimpleStream} delivers consistently strong
performance. With only 4 recent frames, it reaches 67.7\% average
accuracy on OVO-Bench and 80.59\% on StreamingBench. Controlled
ablations further show that the value of longer context is
backbone-dependent rather than uniformly increasing with model scale,
and reveal a consistent perception-memory trade-off: adding more
historical context can improve recall, but often weakens real-time
perception. This suggests that stronger memory, retrieval, or
compression modules should not be taken as evidence of progress unless
they clearly outperform \textsc{SimpleStream} under the same
protocol. We therefore argue that future streaming benchmarks should
separate recent-scene perception from long-range memory, so that performance improvements
from added complexity can be evaluated more clearly.
}

\usepackage{times}
\usepackage{latexsym}
\usepackage[T1]{fontenc}
\usepackage[utf8]{inputenc}
\usepackage{microtype}
\usepackage{graphicx}
\usepackage{booktabs}
\usepackage{tabularx}
\usepackage{multirow}
\usepackage{amsmath}
\usepackage{amssymb}
\usepackage{array}
\usepackage{makecell}
\usepackage{colortbl}
\usepackage{pifont}
\usepackage{caption}
\usepackage{subcaption}
\usepackage{enumitem}
\usepackage{float}
\usepackage{wrapfig}
\usepackage{xcolor}
\usepackage{xspace}
\usepackage{hyperref}
\usepackage{cleveref}

\newcolumntype{L}[1]{>{\raggedright\let\newline\\\arraybackslash\hspace{0pt}}m{#1}}
\newcolumntype{R}[1]{>{\raggedleft\let\newline\\\arraybackslash\hspace{0pt}}m{#1}}

\newcommand{\ignore}[1]{}

\makeatletter
\DeclareRobustCommand\onedot{\futurelet\@let@token\@onedot}
\def\@onedot{\ifx\@let@token.\else.\null\fi\xspace}

\makeatother

\definecolor{MyBlue}{rgb}{0.46, 0.50, 0.61}
\definecolor{MyDarkBlue}{rgb}{0,0.08,0.8}
\definecolor{MyDarkGreen}{RGB}{45,155,45}
\definecolor{MyDarkRed}{rgb}{0.8,0.02,0.02}
\definecolor{MyOrange}{rgb}{1.0, 0.4, 0.2}
\definecolor{MyPurple}{RGB}{111,0,255}
\definecolor{MyRed}{rgb}{0.8,0.0,0.0}
\definecolor{MyGold}{rgb}{0.75,0.6,0.12}
\definecolor{MyDarkgray}{rgb}{0.66, 0.66, 0.66}
\definecolor{MyBrown}{rgb}{0.65, 0.16, 0.16}
\definecolor{MyMutedRose}{rgb}{0.58, 0.29, 0.35}
\definecolor{JiayuanColor}{rgb}{0.60,0.43,0.48}
\definecolor{erranColor}{rgb}{24, 40, 113}

\definecolor{citecolor}{HTML}{696FAD}

\definecolor{bggray}{HTML}{F5F5F5}
\definecolor{pvdblue}{HTML}{DAE8FC}
\definecolor{RoseQuartzBg}{HTML}{F7CAC9}
\definecolor{RoseQuartz}{HTML}{F5A798}
\definecolor{Serenity}{HTML}{92A8D1}
\definecolor{OrangeRed}{rgb}{1.0, 0.27, 0.0}
\definecolor{RoyalBlue}{cmyk}{1, 0.50, 0, 0}
\definecolor{Turquoise}{HTML}{0F4C81}
\definecolor{mint}{rgb}{0.24, 0.71, 0.54}
\definecolor{green}{rgb}{0.0, 0.120, 0.0}

\newdimen\abovecrulesep
\newdimen\belowcrulesep
\makeatletter
\patchcmd{\@@@cmidrule}{\aboverulesep}{\abovecrulesep}{}{}
\patchcmd{\@xcmidrule}{\belowrulesep}{\belowcrulesep}{}{}
\makeatother

\definecolor{mybluetitle}{HTML}{4B527E} %

\definecolor{codegreen}{HTML}{478058}%
\definecolor{codegray}{rgb}{0.5,0.5,0.5}
\definecolor{codepurple}{HTML}{4F5E80} %
\definecolor{backcolour}{rgb}{0.95,0.95,0.92}
\lstdefinestyle{mystyle}{
    backgroundcolor=\color{backcolour},
    commentstyle=\color{codegreen},
    keywordstyle=\color{magenta},
    numberstyle=\tiny\color{codegray},
    stringstyle=\color{codepurple},
    basicstyle=\ttfamily\scriptsize,
    breakatwhitespace=false,
    breaklines=true,
    captionpos=b,
    keepspaces=true,
    frame=none,
    numbersep=5pt,
    showspaces=false,
    showstringspaces=false,
    showtabs=false,
    tabsize=2
}

\newtcolorbox{promptbox}[2][]{
    enhanced, 
    breakable,
    center title,
    left*=0pt, right*=0pt,
    boxsep=2pt, left=5pt, right=5pt,
    skin first=enhanced,
    skin middle=enhanced,
    skin last=enhanced,
    colback  = backcolour,
    fonttitle=\bfseries\rmfamily,
    fontupper=\scriptsize,
    title={\footnotesize\strut{#2}},
    #1
    }

\newtcolorbox{onebox}[2][]{
    enhanced, 
    center title,
    left*=0pt, right*=0pt,
    boxsep=2pt, left=5pt, right=5pt,
    skin first=enhanced,
    skin middle=enhanced,
    skin last=enhanced,
    colframe = mybluetitle!90,
  colback  = mybluetitle!10,
    fonttitle=\bfseries\rmfamily\fontfamily{phv}\selectfont,
    title={\footnotesize\strut{#2}  \refstepcounter{subsubsection} \addcontentsline{toc}{subsubsection}{\string\numberline{\thesubsubsection}#2}
    },
    #1
    }

\usepackage{amsmath,amsfonts,bm}

\def\eqref#1{equation~\ref{#1}}

\def\1{\bm{1}}

\DeclareMathAlphabet{\mathsfit}{\encodingdefault}{\sfdefault}{m}{sl}
\SetMathAlphabet{\mathsfit}{bold}{\encodingdefault}{\sfdefault}{bx}{n}

\providecommand{\citep}[1]{\cite{#1}}
\providecommand{\citet}[1]{\cite{#1}}

\begin{document}

\maketitle

\section{Introduction}
\label{sec:intro}

Streaming video understanding increasingly relies on complex
memory-centric designs to handle long streams under causal
constraints~\citep{Qian2024StreamingLong, Qian2025Dispider, Di2025StreamingVideo, Chen2025Livecc, Yao2025Timechat, Zhang2025Flash, Zeng2025Streamforest, Zhang2026Hermes}.
Across these methods, the complexity typically lies in how past
context is managed, for example through explicit memory banks,
retrieval over prior observations, or compression of visual and
latent representations under bounded budgets. This trend reflects a common assumption: strong streaming
performance requires increasingly complex memory mechanisms.

\begin{figure*}[t]
  \centering
  \includegraphics[width=\linewidth]{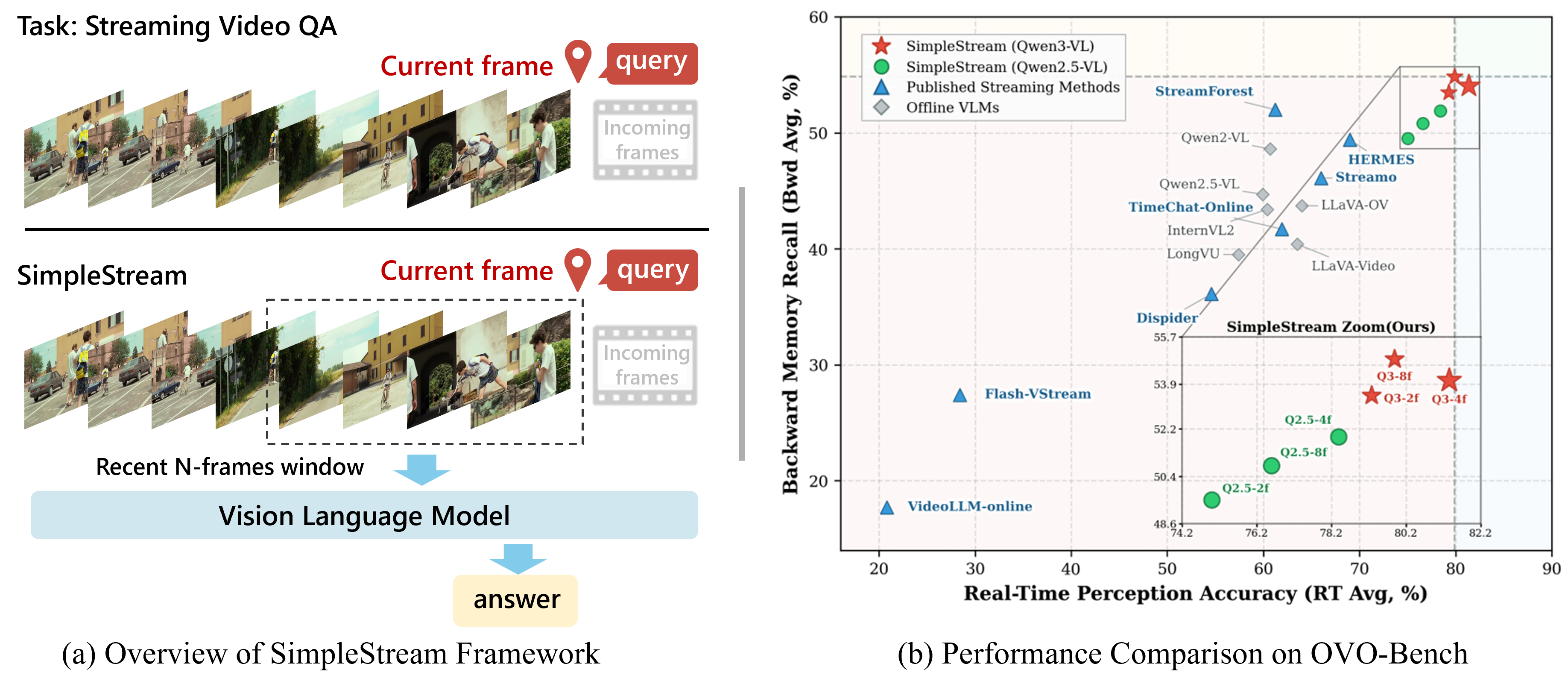}
  \caption{\textbf{A Strong Simple Baseline for Streaming Video Understanding.} (a) Overview of \textsc{SimpleStream}: given a streaming video and a query, only the most recent $N$ frames are fed to an off-the-shelf VLM. (b) Perception-memory comparison on OVO-Bench, where \textsc{SimpleStream} consistently lies on the upper-right frontier across backbone families and window sizes.}
  \label{fig:teaser}
\end{figure*}
However, these increasingly elaborate designs have often delivered
modest or uneven gains, leaving a more fundamental question
underexplored: is such memory complexity actually necessary for
strong streaming video understanding? We show that a very simple
baseline can already outperform published streaming methods.
Figure~\ref{fig:teaser} previews this result, where a minimal
recent-context method lies on the upper-right frontier of both
perception and memory performance.

Motivated by this observation, we introduce \textsc{SimpleStream}, an
intentionally simple baseline for streaming video understanding.
Given a query at time $t$, \textsc{SimpleStream} feeds the last $N$
observed frames and the query text directly to the base VLM.
The design is minimal by construction: preserve only a short recent
window and let a strong backbone operate on clear, uncompressed recent
evidence.

Despite its minimal design, \textsc{SimpleStream} achieves state-of-the-art performance on both OVO-Bench~\citep{Li2025Ovo} and
StreamingBench~\citep{Lin2024Streamingbench}, while also maintaining
the lowest peak GPU memory and competitive latency among all compared
streaming methods. Therefore, a strong backbone plus uncompressed recent
visual context is already a highly competitive streaming solution.
It also exposes two broader issues: injecting additional memory can
degrade real-time perception, and longer context is not uniformly
rewarded even across model scales and backbone families.
These observations make it necessary to report strong simple baselines before
claiming gains from additional streaming complexity.

Our analyses and discussion show that this pattern is systematic rather
than incidental. Across controlled recency-window, model-scaling, and
Visual-RAG ablations, adding more historical context is not uniformly
beneficial. A modest increase in recent context can help, but the
preferred window size depends on model scale and backbone family rather
than increasing monotonically with parameter count. More broadly, extra
historical context can improve memory-oriented behavior, but often at a
cost to present-scene perception. We further show that current benchmark
design can amplify this tension, making aggregate gains difficult to
interpret without separating perception-oriented and memory-oriented
behavior.

These findings motivate a practical evaluation standard for streaming
video understanding: under matched backbones and protocols, reported
gains should be assessed against strong recency baselines and
disaggregated perception-versus-memory metrics, so that the benefit of
added complexity is identifiable rather than assumed.

Our contributions are as follows:
\begin{itemize}
  \item We introduce \textsc{SimpleStream}, a deliberately minimal
  streaming baseline that answers each query using only the last $N$
  frames from the causal prefix with an off-the-shelf VLM, without
  additional memory, retrieval, compression, or training.
  \item Through comprehensive evaluation on OVO-Bench and StreamingBench, we show that
  a simple recent-context baseline is sufficient to outperform prior
  streaming methods while maintaining favorable peak GPU memory and
  latency.
  \item We provide controlled analyses of recent-window size, model
  scale, Visual-RAG augmentation, and benchmark structure, showing that
  the utility of longer context is backbone-dependent rather than
  monotonically improved by larger models, and that memory-side gains
  often come with perception costs.
  \item We argue that future streaming video understanding work should adopt strong recency baselines and disaggregated reporting as default evaluation practice, so that gains from added streaming complexity are demonstrated rather than assumed.
\end{itemize}

\section{Related Work}
\label{sec:related}

\noindent\textbf{Streaming video understanding.}
Recent streaming video understanding research can be broadly organized into three directions: proactive response and interaction, streaming-oriented training, and memory-centric context management. These directions address different challenges in online video understanding, including when a model should respond, how it can align perception with generation under causal constraints, and how it can preserve useful past information under limited compute and memory budgets. Proactive systems concentrate on response timing and interaction policy, for example by predicting answer readiness~\citep{Azad2026Streamready,Liu2026Thinkstream,Xia2025StreamingVideo,Yang2025Livestar}, decoupling decision from
perception~\citep{Qian2025Dispider}, or using an external trigger for response generation~\citep{Wang2025Streambridge}.
Training-oriented approaches instead make online generation feasible
through supervision, positional design, and temporal alignment, as in
LiveCC and Streamo~\citep{Chen2025Livecc,Xia2025StreamingVideo}, without
making memory design the primary object of study.

Memory and context management remain a central concern in streaming video understanding. What differs across methods is not whether history matters, but how explicitly it is modeled and through which mechanism.
Some methods compress or prune redundant tokens or KV states~\citep{Yao2025Timechat,Chen2026StreamingTOM,Jin2025Streamingassistant,Zhang2026Hermes,Wang2026STC}.
Others make past information query-addressable through retrieval or adaptive KV
selection~\citep{Ning2025LiveVLM,Di2025StreamingVideo,Yang2025Streammem}.
Another line introduces explicit external or hierarchical memory
structures~\citep{Xiong2025StreamingVideo,Azad2026Streamready,Guo2026Eventvstream,Zhang2025Flash,Zhou2024StreamingDense,Liu2026Thinkstream},
including StreamForest and FluxMem~\citep{Zeng2025Streamforest,Xie2026Fluxmem}. Some approaches instead summarize long streams through learned latent or
recurrent state, \emph{e.g.}, VideoStreaming~\citep{Qian2024StreamingLong}
and Dispider~\citep{Qian2025Dispider}. Although these methods differ in implementation, they are often motivated by the view that strong streaming performance benefits from complex memory
mechanisms for preserving history. \textsc{SimpleStream} makes the
recent-context baseline the primary reference point. This reframing turns what is usually a peripheral ablation into a direct test of when
additional memory mechanisms deliver meaningful gains.

\noindent\textbf{Streaming video benchmarks.}
Evaluating streaming video understanding requires benchmarks that
distinguish among causal online reasoning, proactive interaction, and
retrospective video comprehension. OVO-Bench~\citep{Li2025Ovo},
StreamingBench~\citep{Lin2024Streamingbench}, and other causal
benchmarks~\citep{Huang2025OnlineVideo,Liu2026Vcbench,Xu2025Streamingvlm,Xia2025StreamingVideo,Zeng2025Streamforest,Jin2025Streamingassistant,Chen2025Livecc}
evaluate models under
observed-only constraints, requiring both current-scene perception and
the use of prior context. Benchmarks such as
OmniMMI~\citep{Wang2025Omnimmi} and
ProactiveVideoQA~\citep{Wang2025Proactivevideoqa} instead emphasize
initiative, assistance, and turn-taking, a direction further developed
by RIVER, LiViBench, and
PhoStream~\citep{Shi2026River,Wang2026Livibench,Lu2026Phostream}, with
additional benchmark formulations~\citep{Azad2026Streamready,Wang2025Streambridge,Zhang2025Eyes}. By
contrast, retrospective or offline video understanding benchmarks such
as LVBench~\citep{Wang2025Lvbench}, MLVU~\citep{Zhou2025Mlvu}, and
EgoLifeQA~\citep{Yang2025Egolife} target long-range temporal reasoning
and event understanding over full videos, but do not impose the same
causal streaming constraint~\citep{Fu2025VideoMme,Li2024Mvbench,Mangalam2023Egoschema}.
To enable a
cleaner comparison between a strong simple baseline and more elaborate
streaming designs, we thus focus primarily on OVO-Bench and use
StreamingBench as a complementary evaluation.

\begin{figure*}[t]
  \centering
  \includegraphics[width=\textwidth]{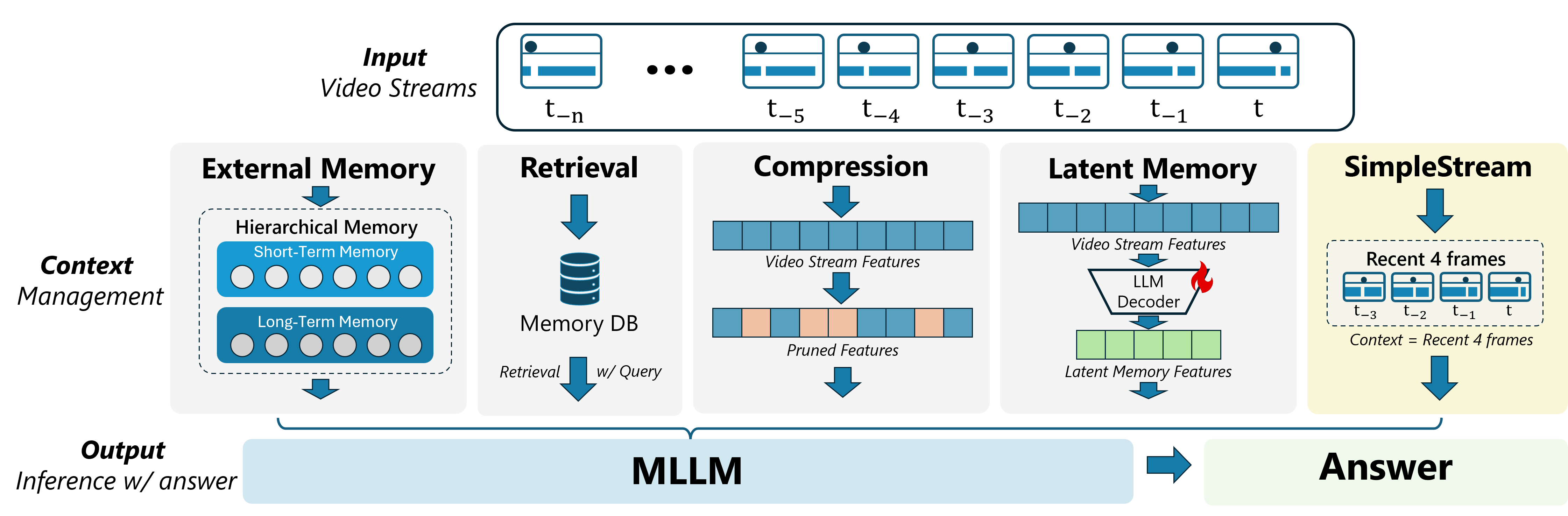}
  \caption{\textbf{A landscape of streaming video understanding methods.} Most existing approaches differ mainly in how they preserve and reuse historical information under bounded budgets, while \textsc{SimpleStream} keeps only a small recent frame window.}
  \label{fig:streaming_landscape}
\end{figure*}

\section{From Complex Streaming Methods to SimpleStream}
\label{sec:method}

\subsection{A Landscape of Streaming Video Understanding Methods}

We study streaming video understanding under a causal observation protocol, where the model must answer each query using only past observations. At query time $t$, the model must answer a text question
$q_t$ using only the observed prefix of the video stream up to $t$,
without access to future frames or side information. Although the
visible prefix may grow arbitrarily long, the model can condition only
on a bounded representation at inference time under realistic limits on
memory, attention tokens, and per-step computation. We therefore view
streaming inference as a context-management problem: at each query, the
system must construct a bounded working context $C_t$ from the observed
history. This context must keep the answer grounded in what has been
seen while respecting a fixed streaming budget. Under this formulation, streaming
QA is not simply sequential decoding over a video prefix, but causal,
budgeted context management.

Figure~\ref{fig:streaming_landscape} groups prior methods by the mechanism
that expands $C_t$ beyond a short recency window. External-memory
systems maintain structured history online: StreamForest~\citep{Zeng2025Streamforest}
maintains an event-level tree whose insertion rule balances an explicit
penalty with temporal distance, content similarity, and merge
frequency, while Flash-VStream~\citep{Zhang2025Flash} uses fixed-size
Flash Memory with slots reserved for global summaries and salient frame
details. Retrieval-based methods retain past representations so they can be selected at query time. ReKV~\citep{Di2025StreamingVideo} stores a disk-backed historical
KV cache and loads query-conditioned KV states when a question arrives.
Compression targets the KV and attention budget directly;
HERMES~\citep{Zhang2026Hermes} maintains a hierarchical memory view over
the KV state by reusing a compact cached representation. Latent-memory approaches learn a
constant-length state for the prefix: VideoStreaming~\citep{Qian2024StreamingLong}
and Dispider~\citep{Qian2025Dispider} use a compact LLM decoder to
compress the observed stream into fixed-size memory features, relying on
learned implicit memory and substantial supervised fine-tuning. Despite their differences, these methods share the goal of expanding or restructuring $C_t$ under a fixed streaming budget. This framing motivates a minimal baseline that constructs $C_t$ only from the most recent observations.

\subsection{SimpleStream: A Simple Recent-N-Frames Baseline}

Unlike prior streaming systems, we do not introduce an additional mechanism for managing long-range history. Instead, we introduce \textsc{SimpleStream}, a deliberately simple
baseline that isolates what current off-the-shelf VLMs can achieve using only recent visual context. Let the video stream be
represented as a sequence of frames, where $f_i$ denotes the visual
frame at time step $i$. Given a question $q_t$ at time $t$, we feed the
base VLM only the most recent $N$ frames and the text query:
\[
\textsc{SimpleStream}(t)= \mathrm{VLM}\bigl(\{f_{t-N+1},\ldots,f_t\},\, q_t\bigr)
\]
By construction, \textsc{SimpleStream} omits the additional memory mechanisms used in prior streaming systems. Frames outside the sliding window are discarded, so the per-query memory
and computation remain bounded by $N$ and do not grow with stream
length. \textsc{SimpleStream} introduces no architectural modification,
memory module, or additional training; it is an inference-time input
policy applied to an off-the-shelf VLM. We use it as a controlled
reference baseline to isolate how much streaming performance can be
obtained from recent visual context alone, while minimizing confounding
effects from additional training, module design, or system-level
engineering.

\begin{table*}[t!]
\centering
\caption{\textbf{Main results on OVO-Bench~\citep{Li2025Ovo} and
StreamingBench~\citep{Lin2024Streamingbench}.}
Per-task accuracy on OVO-Bench under its Real-Time Visual Perception and
Backward Tracing tracks.
``StreamingBench'' denotes RTVU accuracy.
``--'' indicates unreported results.
$\dagger$: Qwen2.5-VL-7B + HERMES~(4K tokens).
Task abbreviations follow OVO-Bench: OCR, ACR, ATR, STU, FPD, and OJR denote Optical Character Recognition, Action Recognition, Attribute Recognition, Spatial Understanding, Future Prediction, and Object Recognition; EPM, ASI, and HLD denote Episodic Memory, Action Sequence Identification, and Hallucination Detection.
``Avg.'' is the mean of Real-Time and Backward category averages.
For a fair comparison with most baselines, we sample at 1\,fps in \textsc{SimpleStream} and cap maximum frames at 2, 4, or 8.}
\label{tab:main_ovo}
\label{tab:main_streaming}
\newcommand{\best}[1]{\begingroup\setlength{\fboxsep}{1pt}\colorbox{yellow!35}{\textbf{#1}}\endgroup}
\renewcommand{\arraystretch}{1.1}
\setlength{\tabcolsep}{3pt}
\resizebox{\textwidth}{!}{%
\begin{tabular}{lc|c|cccccc|c|ccc|c|c}
\toprule
\multirow{3}{*}{Model} & \multirow{3}{*}{\#Frames}
& \multirow{3}{*}{\shortstack{Streaming\\Bench}}
& \multicolumn{12}{c}{\textit{OVO-Bench}} \\
\cmidrule(lr){4-15}
& &
& \multicolumn{7}{c|}{\textit{Real-Time Visual Perception}}
& \multicolumn{4}{c|}{\textit{Backward Tracing}}
& \multirow{2}{*}{Avg.} \\
\cmidrule(lr){4-10} \cmidrule(lr){11-14}
& & & OCR & ACR & ATR & STU & FPD & OJR & Avg.
& EPM & ASI & HLD & Avg. & \\
\midrule
Human
& -- & 91.46
& 94.0 & 92.6 & 94.8 & 92.7 & 91.1 & 94.0 & 93.2
& 92.6 & 93.0 & 91.4 & 92.3
& 92.77 \\
\midrule
\multicolumn{15}{c}{\textit{Offline Video LLMs}} \\
\midrule
Qwen2.5-VL-7B~\citep{Bai2025Qwen25VL}
& 1\,fps & 73.31
& 67.8 & 55.1 & 67.2 & 42.1 & 66.3 & 60.9 & 59.9
& 51.5 & 58.8 & 23.7 & 44.7
& 52.28 \\
LLaVA-OneVision-7B~\citep{Li2024LLaVAOneVision}
& 32 & 71.12
& 66.4 & 57.8 & 73.3 & 53.4 & 71.3 & 62.0 & 64.0
& 54.2 & 55.4 & 21.5 & 43.7
& 53.85 \\
InternVL2-8B~\citep{Chen2024InternVL2}
& 16 & 63.72
& 67.1 & 60.6 & 63.8 & 46.1 & 68.3 & 56.5 & 60.4
& 48.2 & 57.4 & 24.7 & 43.4
& 51.90 \\
LLaVA-Video-7B~\citep{Zhang2024LLaVAVideo}
& 64 & --
& 69.1 & 58.7 & 68.8 & 49.4 & 74.3 & 59.8 & 63.5
& \underline{56.2} & 57.4 &  7.5 & 40.4
& 51.95 \\
Qwen2-VL-7B~\citep{Wang2024Qwen2VL}
& 64 & 69.04
& 69.1 & 53.2 & 63.8 & 50.6 & 66.3 & 60.9 & 60.7
& 44.4 & \best{66.9} & 34.4 & 48.6
& 54.62 \\
LongVU-7B~\citep{Shen2024LongVU}
& 1\,fps & --
& 55.7 & 49.5 & 59.5 & 48.3 & 68.3 & 63.0 & 57.4
& 43.1 & \underline{66.2} &  9.1 & 39.5
& 48.45 \\
\midrule
\multicolumn{15}{c}{\textit{Online / Streaming Video LLMs}} \\
\midrule
VideoLLM-online-8B~\citep{Wang2025Videollm}
& 2\,fps & 35.99
&  8.1 & 23.9 & 12.1 & 14.0 & 45.5 & 21.2 & 20.8
& 22.2 & 18.8 & 12.2 & 17.7
& 19.26 \\
Flash-VStream-7B~\citep{Zhang2025Flash}
& 1\,fps & 23.23
& 24.2 & 29.4 & 28.5 & 33.7 & 25.7 & 28.8 & 28.4
& 39.1 & 37.2 &  5.9 & 27.4
& 27.90 \\
Dispider-7B~\citep{Qian2025Dispider}
& 1\,fps & 67.63
& 57.7 & 49.5 & 62.1 & 44.9 & 61.4 & 51.6 & 54.6
& 48.5 & 55.4 &  4.3 & 36.1
& 45.35 \\
TimeChat-Online-7B~\citep{Yao2025Timechat}
& 1\,fps & 75.28
& 75.2 & 46.8 & 70.7 & 47.8 & 69.3 & 61.4 & 61.9
& 55.9 & 59.5 &  9.7 & 41.7
& 51.80 \\
StreamForest-7B~\citep{Zeng2025Streamforest}
& 1\,fps & 77.26
& 68.5 & 53.2 & 71.6 & 47.8 & 65.4 & 60.9 & 61.2
& \best{58.9} & 64.9 & 32.3 & 52.0
& 56.60 \\
Streamo-7B~\citep{Xia2025StreamingVideo}
& 1\,fps & --
& 79.2 & 57.8 & 75.0 & 49.4 & 64.4 & 70.1 & 66.0
& 54.6 & 52.0 & 31.7 & 46.1
& 56.05 \\
HERMES-7B$^\dagger$~\citep{Zhang2026Hermes}
& 1\,fps & \underline{79.44}
& 85.2 & 64.2 & 71.6 & 53.4 & 74.3 & 65.2 & 69.0
& 48.5 & 62.2 & 37.6 & 49.4
& 59.20 \\
\midrule
\multicolumn{15}{c}{\textit{\textsc{SimpleStream} (Ours)}} \\
\midrule
Qwen2.5-VL-7B + 2f
& 2 & 76.39
& 88.6 & 67.0 & 81.0 & 64.6 & 69.3 & 79.3 & 75.0
& 49.2 & 56.8 & 42.5 & 49.5
& 62.22 \\
Qwen2.5-VL-7B + 4f
& 4 & 78.47
& 94.0 & 72.5 & 80.2 & 68.0 & 76.2 & 79.3 & 78.4
& 54.5 & 60.8 & 40.3 & 51.9
& 65.13 \\
Qwen2.5-VL-7B + 8f
& 8 & \underline{79.11}
& \best{95.3} & 67.9 & 79.3 & 61.2 & 74.3 & \underline{81.5} & 76.6
& 52.2 & 63.5 & 36.6 & 50.8
& 63.70 \\
Qwen3-VL-8B + 2f
& 2 & 78.31
& 89.3 & 77.1 & \best{83.6} & \best{68.5} & 76.2 & 81.0 & 79.3
& 49.5 & 56.1 & \best{54.8} & 53.5
& 66.38 \\
Qwen3-VL-8B + 4f
& 4 & \best{80.59}
& \underline{94.0} & \best{85.3} & \underline{82.8} & 65.7 & \best{77.2} & \best{83.2} & \best{81.4}
& 51.9 & 58.1 & \underline{52.1} & \underline{54.0}
& \best{67.70} \\
Qwen3-VL-8B + 8f
& 8 & 78.83
& \underline{94.0} & \underline{84.4} & 80.2 & 64.0 & 75.3 & \underline{81.5} & \underline{79.9}
& 53.2 & 60.8 & 50.5 & \best{54.9}
& \underline{67.37} \\
\bottomrule
\end{tabular}
}
\end{table*}

\section{Experiments}
\label{sec:experiments}

\subsection{Experimental Setup}
\label{sec:setup}

\paragraph{Benchmarks.}
We evaluate all models on OVO-Bench~\citep{Li2025Ovo} and
StreamingBench~\citep{Lin2024Streamingbench}.
OVO-Bench contains 1,640 questions over 12 tasks spanning memory recall,
real-time perception, and future-oriented reasoning. We evaluate the
Backward Tracing and Real-Time Visual Perception categories, which directly test the trade-off between
memory and real-time perception. Under the official evaluation protocol,
each question is answered using only the video prefix available up to the
query time, and we report scores with the official scorer. Offline and
online baselines therefore share the same visible prefix at query time,
while retaining their own model-specific inference pipelines, prompts,
and frame budgets or frame rates as specified in the original papers or
official implementations. For StreamingBench, we use the official
real-time visual understanding subset, which contains 2,500 questions
across ten task types. This setting complements OVO-Bench with broader
real-time coverage and tests whether the same trends transfer across
benchmarks.

\paragraph{Compared Models.}
We compare against six offline video LLMs~\citep{Bai2025Qwen25VL,Li2024LLaVAOneVision,Chen2024InternVL2,Zhang2024LLaVAVideo,Wang2024Qwen2VL,Shen2024LongVU}
and seven representative streaming video LLMs~\citep{Wang2025Videollm,Zhang2025Flash,Qian2025Dispider,Yao2025Timechat,Zeng2025Streamforest,Xia2025StreamingVideo,Zhang2026Hermes},
covering the main design paradigms in recent streaming video understanding; the full model list is reported in Table~\ref{tab:main_ovo}. Unless otherwise noted, we use the best inference settings reported in the original papers or official implementations, and summarize each model's frame budget or sampling setting in the \#Frames column. For special cases, we follow the table notes, e.g., HERMES$^\dagger$ denotes the Qwen2.5-VL-7B variant with a 4K-token memory budget.

\paragraph{Our \textsc{SimpleStream}.}
We instantiate \textsc{SimpleStream} with two open-source VLM backbones,
Qwen2.5-VL-7B-Instruct~\citep{Bai2025Qwen25VL} and
Qwen3-VL-8B-Instruct~\citep{Bai2025Qwen3VL}. At each query, we sample the
visible stream at 1\,fps and feed the model only the last
$N\in\{2,4,8\}$ frames. By default, \textsc{SimpleStream} uses no
separate memory bank, retrieval, or vision/KV compression beyond this
recent window.

\subsection{Benchmark Performance}
\label{sec:main_results}

Table~\ref{tab:main_ovo} reports our main results on OVO-Bench and
StreamingBench, comparing offline VLMs, online/streaming VLMs, and
\textsc{SimpleStream} under a unified protocol.


On OVO-Bench, the best \textsc{SimpleStream} configuration (Qwen3-VL, 4
frames) reaches 67.7\%, exceeding the strongest published streaming method,
HERMES~\citep{Zhang2026Hermes}, by 8.5\,pp (59.2\%). The gain is largest on Real-Time Visual
Perception, where \textsc{SimpleStream} achieves 81.4\% versus 69.0\% for
HERMES, with especially large margins on the OCR, ACR, and OJR tracks. On Backward
Tracing, \textsc{SimpleStream} remains competitive: the 8-frame variant
reaches 54.9\%, compared with 52.0\% for StreamForest~\citep{Zeng2025Streamforest} and 49.4\% for HERMES. The same pattern appears on StreamingBench. \textsc{SimpleStream} with
Qwen3-VL and 4 frames reaches 80.59\%, surpassing HERMES
(79.44\%), while five of the six \textsc{SimpleStream} configurations
outperform StreamForest, and all six outperform the remaining streaming
baselines other than HERMES.

\begin{table*}[t]
\centering
\caption{\textbf{Model scale effects under a fixed recent-window evaluation
protocol on OVO-Bench.}
We vary model scale within each backbone family while keeping all other
evaluation settings unchanged. For each recent window, we report Backward
Tracing accuracy, Real-Time Visual Perception accuracy, and their mean
(``Avg.''). The 7B and 8B rows reuse the corresponding 2-frame, 4-frame, and
8-frame runs from the main experiment under the same protocol. ``--''
indicates unavailable or incomplete runs.}
\label{tab:model_scale_effects}
\small
\setlength{\tabcolsep}{3.5pt}
\resizebox{\textwidth}{!}{%
\begin{tabular}{l|ccc|ccc|ccc|ccc}
\toprule
\multirow{2}{*}{Model}
& \multicolumn{3}{c|}{2 Frames}
& \multicolumn{3}{c|}{4 Frames}
& \multicolumn{3}{c|}{8 Frames}
& \multicolumn{3}{c}{16 Frames} \\
\cmidrule(lr){2-4}
\cmidrule(lr){5-7}
\cmidrule(lr){8-10}
\cmidrule(lr){11-13}
& Bwd. & Real-Time & Avg.
& Bwd. & Real-Time & Avg.
& Bwd. & Real-Time & Avg.
& Bwd. & Real-Time & Avg. \\
\midrule
\multicolumn{13}{c}{\textit{Qwen2.5-VL}} \\
\midrule
Qwen2.5-VL-3B  & 42.09 & 70.87 & 56.48 & 42.89 & 73.47 & \textbf{58.18} & 40.98 & 72.07 & 56.52 & 45.75 & 70.60 & 58.17 \\
Qwen2.5-VL-7B  & 49.50 & 75.00 & 62.22 & 51.90 & 78.40 & \textbf{65.13} & 50.80 & 76.60 & 63.70 & 48.87 & 74.45 & 61.66 \\
Qwen2.5-VL-32B & 45.87 & 78.80 & 62.33 & 48.64 & 83.03 & \textbf{65.84} & 49.43 & 81.70 & 65.56 & 49.97 & 80.69 & 65.33 \\
Qwen2.5-VL-72B & 57.65 & 77.63 & 67.64 & 58.71 & 80.36 & 69.53 & 56.75 & 81.25 & 69.00 & 60.61 & 80.91 & \textbf{70.76} \\
\midrule
\multicolumn{13}{c}{\textit{Qwen3-VL}} \\
\midrule
Qwen3-VL-2B  & 43.92 & 73.17 & 58.55 & 45.00 & 76.07 & \textbf{60.53} & 45.00 & 75.25 & 60.12 & 47.41 & 73.29 & 60.35 \\
Qwen3-VL-4B  & 52.01 & 76.04 & 64.03 & 52.37 & 79.25 & 65.81 & 53.28 & 78.67 & 65.97 & 54.63 & 77.48 & \textbf{66.06} \\
Qwen3-VL-8B  & 53.50 & 79.30 & 66.38 & 54.00 & 81.40 & \textbf{67.70} & 54.90 & 79.90 & 67.37 & 56.41 & 77.88 & 67.15 \\
Qwen3-VL-32B & 63.21 & 81.65 & 72.43 & 63.41 & 83.57 & 73.49 & 65.39 & 82.78 & \textbf{74.09} & 66.93 & 80.69 & 73.81 \\
Qwen3-VL-30B-A3B & 58.59 & 82.69 & 70.64 & 61.00 & 85.56 & \textbf{73.28} & 61.11 & 84.59 & 72.85 & 64.08 & 81.85 & 72.97 \\
\bottomrule
\end{tabular}%
}
\end{table*}

\begin{figure}[H]
  \centering
  \begin{minipage}[c]{0.46\textwidth}
    \centering
    \includegraphics[width=\linewidth]{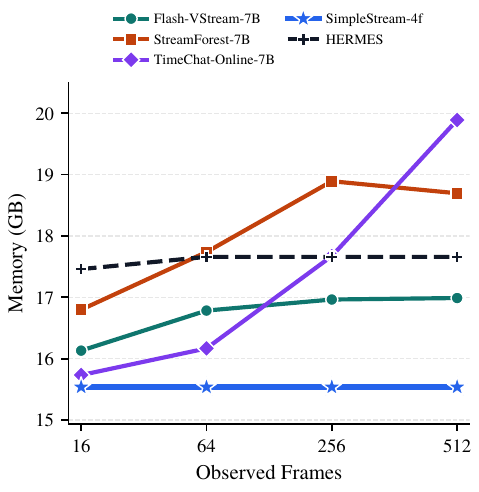}
  \end{minipage}%
  \hfill
  \begin{minipage}[c]{0.50\textwidth}
    \centering
    \small
\setlength{\tabcolsep}{4pt}
\resizebox{\linewidth}{!}{%
\begin{tabular}{@{}lccc@{}}
\toprule
Method & 16 Frames & 64 Frames & 256 Frames \\
\midrule
Dispider~\citep{Qian2025Dispider}
            & 490 & 1460 & 3810 \\
ReKV~\citep{Di2025StreamingVideo}
                & 250 & 380  & 740  \\
LiveVLM~\citep{Ning2025LiveVLM}
             & 240 & 310  & 600  \\
StreamForest~\citep{Zeng2025Streamforest}
        & 221 & 560  & 834  \\
StreamingTOM~\citep{Chen2026StreamingTOM}
        & 180 & 180  & 280  \\
TimeChat-Online~\citep{Yao2025Timechat}
     & 156 & 607  & 3072 \\
Flash-VStream~\citep{Zhang2025Flash}
       & 36  & 63   & 67   \\
HERMES~\citep{Zhang2026Hermes}
              & \textbf{27} & \textbf{29} & \textbf{29} \\
\midrule
\multicolumn{4}{c}{\textit{Our Method}} \\
\midrule
\textsc{SimpleStream}-4f & \underline{35} & \underline{33} & \underline{38} \\
\bottomrule
\end{tabular}%
}

  \end{minipage}

  \vspace{0.3cm}

  \begin{minipage}[t]{0.46\textwidth}
    \captionof{figure}{\textbf{Peak GPU memory vs.\ observed frames.}
    \textsc{SimpleStream}-4f maintains the lowest and flattest memory curve because it retains only a fixed recent frame window.}
    \label{fig:vram_scaling_sota}
  \end{minipage}%
  \hfill
  \begin{minipage}[t]{0.50\textwidth}
    \captionof{table}{\textbf{Latency-efficient streaming inference (TTFT).}
    TTFT (ms) for streaming baselines at 16, 64, and 256 observed frames.
    SimpleStream-4f uses a Qwen2.5-VL-7B backbone and remains close to the fastest method.}
    \label{tab:ttft_comparison}
  \end{minipage}
\end{figure}

\subsection{Model Scale Effects}

In the main experiment, increasing the recent window from 4 to 8 frames does
not consistently improve performance, even though the longer window strictly
contains the shorter one. To examine whether this non-monotonic behavior
depends on model scale, we extend the study under the same OVO-Bench protocol
to multiple sizes of Qwen2.5-VL and Qwen3-VL, evaluating \textsc{SimpleStream}
with recent windows $N\in\{2,4,8,16\}$ across all publicly available
Qwen2.5-VL checkpoints up to 72B and all publicly available Qwen3-VL
checkpoints up to 32B, plus an additional Qwen3-VL-30B-A3B checkpoint, while
keeping all other settings fixed.

Table~\ref{tab:model_scale_effects} reports the detailed results. Across both
backbone families, moving from 2 to 4 frames usually improves average
accuracy. For many small and mid-sized checkpoints, performance then plateaus
or slightly declines as the window expands further. Larger windows can become
more favorable for some higher-capacity checkpoints, but the preferred window
size varies across scales and backbone families. Overall, these results show
that model scale affects the optimal recent-window size, without changing the
main conclusion that longer context is not uniformly better.

\subsection{Efficiency Observations}
\label{sec:efficiency}

We also evaluated and compared the efficiency of the models, including time to
first token (TTFT) and peak GPU memory.
Table~\ref{tab:ttft_comparison} shows that \textsc{SimpleStream}-4f remains
latency-competitive despite using no explicit memory module. Across
increasing observed-frame budgets, it achieves lower TTFT than most published
streaming baselines. HERMES~\citep{Zhang2026Hermes} is the only method that is
consistently faster. The remaining gap is modest despite
\textsc{SimpleStream}-4f using no dedicated memory module. This pattern
suggests that low startup latency does not require persistent historical
state. \textsc{SimpleStream}-4f attains the second-lowest TTFT at 16, 64, and
256 observed frames. Figure~\ref{fig:vram_scaling_sota} complements the
latency comparison by showing that \textsc{SimpleStream}-4f also has the
lowest peak GPU memory usage. Unlike external-memory streaming systems, its
state size does not accumulate with the observed stream. This behavior follows
directly from the minimalist design in Section~\ref{sec:method}: the model
preserves only a fixed recent frame window, so memory usage remains nearly
flat as the stream grows.

\begin{figure}[!t]
  \centering
  \begin{minipage}[c]{0.46\textwidth}
    \centering
    \includegraphics[width=\linewidth]{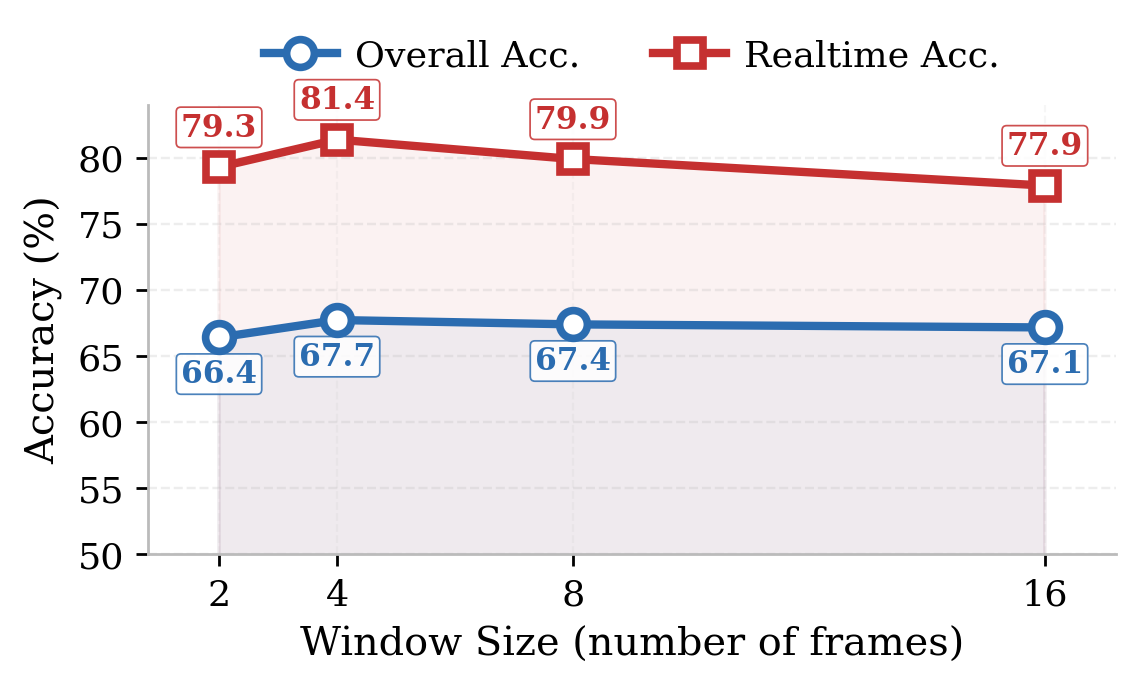}
  \end{minipage}%
  \hfill
  \begin{minipage}[c]{0.50\textwidth}
    \centering
    \footnotesize
\resizebox{\linewidth}{!}{%
\begin{tabular}{@{}lccr@{}}
\toprule
Track & Base & +V-RAG & $\Delta$\,Acc. \\
\midrule
\multicolumn{4}{c}{\textit{Backward Tracing}} \\
\cmidrule(lr){1-4}
EPM {\scriptsize(Episodic Memory)}                & 52.5 & 59.6 & \textcolor{green!60!black}{$+7.1$} \\
ASI {\scriptsize(Action Sequence Identification)}  & 58.8 & 64.9 & \textcolor{green!60!black}{$+6.1$} \\
HLD {\scriptsize(Hallucination Detection)}         & 45.7 & 33.3 & \textcolor{red!70!black}{$-12.4$} \\
\midrule
\multicolumn{4}{c}{\textit{Real-Time Visual Perception}} \\
\cmidrule(lr){1-4}
OJR {\scriptsize(Object Recognition)}              & 81.5 & 72.3 & \textcolor{red!70!black}{$-9.2$} \\
ATR {\scriptsize(Attribute Recognition)}           & 81.9 & 81.9 & $0.0$ \\
ACR {\scriptsize(Action Recognition)}              & 78.9 & 71.6 & \textcolor{red!70!black}{$-7.3$} \\
OCR {\scriptsize(Optical Character Recognition)}   & 94.0 & 85.9 & \textcolor{red!70!black}{$-8.1$} \\
FPD {\scriptsize(Future Prediction)}               & 77.2 & 74.3 & \textcolor{red!70!black}{$-2.9$} \\
STU {\scriptsize(Spatial Understanding)}           & 64.0 & 62.4 & \textcolor{red!70!black}{$-1.6$} \\
\midrule
Acc.                                               & 66.0 & 63.7 & \textcolor{red!70!black}{$-2.3$} \\
\bottomrule
\end{tabular}%
}

  \end{minipage}

  \vspace{0.3cm}

  \begin{minipage}[t]{0.46\textwidth}
    \captionof{figure}{\textbf{Window-size ablation.} Under this controlled setting, \textsc{SimpleStream} reaches its highest Real-Time accuracy with 4 recent frames, while overall accuracy does not improve monotonically as the window widens.}
    \label{fig:window_size_accuracy}
  \end{minipage}%
  \hfill
  \begin{minipage}[t]{0.50\textwidth}
    \captionof{table}{\textbf{Per-track Visual-RAG analysis on OVO-Bench.}
    Base: matched recent-window baseline; +V-RAG: 5 retrieved chunks appended.
    Acc.\ $=(\mathrm{RT}+\mathrm{Bwd})/2$.}
    \label{tab:rag_tradeoff}
  \end{minipage}
\end{figure}

\section{Analysis}
\label{sec:perceptual_fidelity}

\subsection{Longer Context Is Not Always Better}
\label{sec:window_size}

A common assumption in streaming video understanding is that giving the model
more past visual evidence should improve answers, because many decisions depend
on continuity across time and earlier observations.
We test that assumption with three complementary probes.
First, we enlarge the recent-frame window fed to the VLM to measure the basic
effect of exposing more contiguous visual context.
Second, we analyze model scaling to test whether the utility of longer visual
context depends on backbone capacity.
Third, we append retrieved historical chunks with Visual-RAG to examine whether
selectively re-injecting distant frames changes the conclusion.
Taken together, these analyses turn a narrow ablation result into a broader
practical question: more context is not free, and whether it helps depends on
whether the backbone can actually absorb and use it.

\paragraph{Recency-window ablation.}
The simplest way to extend context is to widen the recent-frame window itself.
Holding retrieval, prompting, and decoding otherwise fixed, we vary only the
number of consecutive frames shown to the VLM.
Figure~\ref{fig:window_size_accuracy} reports a controlled ablation over
$N \in \{2,4,8,16\}$ recent frames.
Moving from 2 to 4 frames improves both Overall accuracy (66.4 $\rightarrow$ 67.7)
and Real-Time accuracy (79.3 $\rightarrow$ 81.4), which indicates that a modestly
wider recent view still supplies useful temporal cues.
Beyond this point, however, performance does not keep rising: at 8 frames,
Overall falls to 67.4 and Real-Time accuracy to 79.9, and at 16 frames they decline further
to 67.1 and 77.9.
Taken together, accuracy is non-monotonic in window size: a modest expansion
helps, but further growth yields flat or declining scores, inconsistent with
the expectation that simply stacking more recent frames should monotonically
improve answers.

\FloatBarrier

\begin{figure}[!t]
  \centering
  \includegraphics[width=\linewidth]{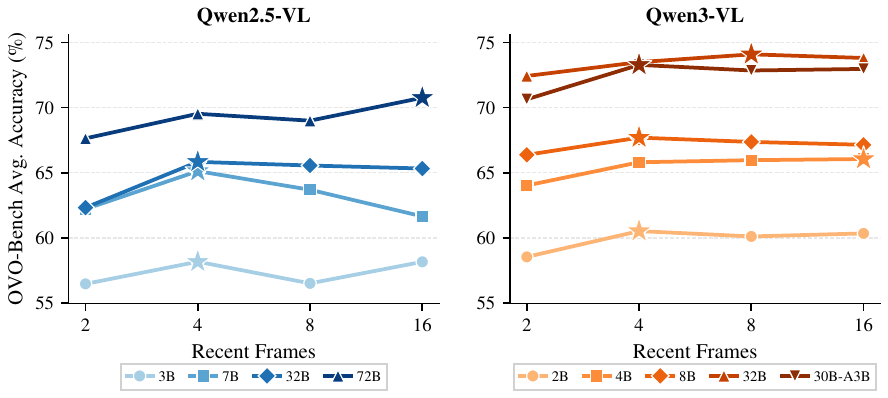}
  \caption{\textbf{Model-scaling ablation on OVO-Bench.}
  Average accuracy versus recent-window size for Qwen2.5-VL (left) and
  Qwen3-VL (right) checkpoints. Stars mark the best window for each
  checkpoint. Many checkpoints peak at 4f, but several prefer longer
  windows, including Qwen3-VL-4B at 16f and larger checkpoints at 8f or
  16f.}
  \label{fig:model_scale_ablation}
\end{figure}

\paragraph{Model-scaling ablation.}
Table~\ref{tab:model_scale_effects} and
Figure~\ref{fig:model_scale_ablation} suggest that this trend is better
understood as a model-scale effect than a clean scaling law, since the performance is non-monotonic in window size and varies across checkpoints. Larger models
sometimes benefit more from wider recent windows, but the optimal window does
not increase uniformly with parameter count. One plausible explanation is that whether a model benefits from additional context depends on its capacity to process it. Longer
windows add potentially useful evidence, but they also introduce more
redundancy. Smaller
models therefore saturate earlier, with a short recent window already capturing
most of the usable signal. Larger models can absorb more of this added context,
which explains why 8f or 16f becomes competitive, and occasionally optimal, only for some higher-capacity checkpoints.

Model scale alone, however, does not determine the optimum. For example, Qwen2.5-VL-72B prefers 16 frames, whereas Qwen2.5-VL-32B
peaks at 4 frames; similarly, Qwen3-VL-32B prefers 8 frames, while
Qwen3-VL-30B-A3B peaks at 4 frames. In addition, OVO-Bench does not uniformly reward wider
recent windows, because many questions are still dominated by
present-scene perception. Taken
together, these results suggest that the effective context range is not
a universally increasing function of model scale, but a quantity shaped
by backbone family and by how the benchmark balances present-scene
perception against the use of historical context.

\FloatBarrier

\paragraph{Visual-RAG ablation.}
Rather than widening the contiguous window, Visual-RAG~\citep{Lewis2020RAG} probes the same
``more history helps'' premise through targeted retrieval.
A CLIP-based~\citep{Radford2021CLIP} index over historical chunks is built offline; at inference time,
the top-5 most similar past chunks are appended to the recent-frame input
before the VLM generates an answer.
Table~\ref{tab:rag_tradeoff} shows that this enrichment is again not uniformly
beneficial. We therefore focus on the within-table deltas between the
matched base and +V-RAG conditions, rather than direct row-by-row
comparison to Table~\ref{tab:main_ovo}.
Visual-RAG improves some Backward tracks, especially EPM (+7.1) and ASI (+6.1),
which confirms that retrieval can recover useful historical evidence, but those
gains coincide with clear degradations on Real-Time tracks, including OJR
(-9.2), OCR (-8.1), and ACR (-7.3), and overall accuracy drops from 66.0 to 63.7.
The outcome is again mixed rather than uniformly positive: selected Backward
tracks rise, yet multiple Real-Time tracks fall and overall accuracy drops, so
richer historical evidence does not translate into reliable aggregate gains
under this setup.

These three probes tell a consistent story.
A slightly longer recent window helps, but gains saturate quickly; beyond that
point, longer context helps only when the backbone has enough capacity to use
it, and even targeted retrieval fails to deliver uniform gains.
Benefits appear task-local, whereas costs often surface on
perception-oriented evaluation slices, which is especially consequential when
aggregate scores overweight Real-Time-style tracks.
This observation motivates the next subsection, where we quantify the trade-off
between historical gains and real-time perception loss.

\subsection{Perception-Memory Trade-off}
The preceding results reveal an asymmetric pattern: adding historical context
often hurts current-scene perception, while its benefits on memory are narrower
and task-dependent. Raw OVO-Bench category averages do not expose this pattern
cleanly because they mix the effect of the streaming mechanism with backbone
strength and input budget. In addition, the Backward split is not a pure memory
measure, since HLD primarily reflects hallucination robustness rather than
episodic event recall. We therefore separate the two sides of the trade-off.

\begin{wrapfigure}{r}{0.48\columnwidth}
  \vspace{-0.9\baselineskip}
  \centering
  \includegraphics[width=\linewidth]{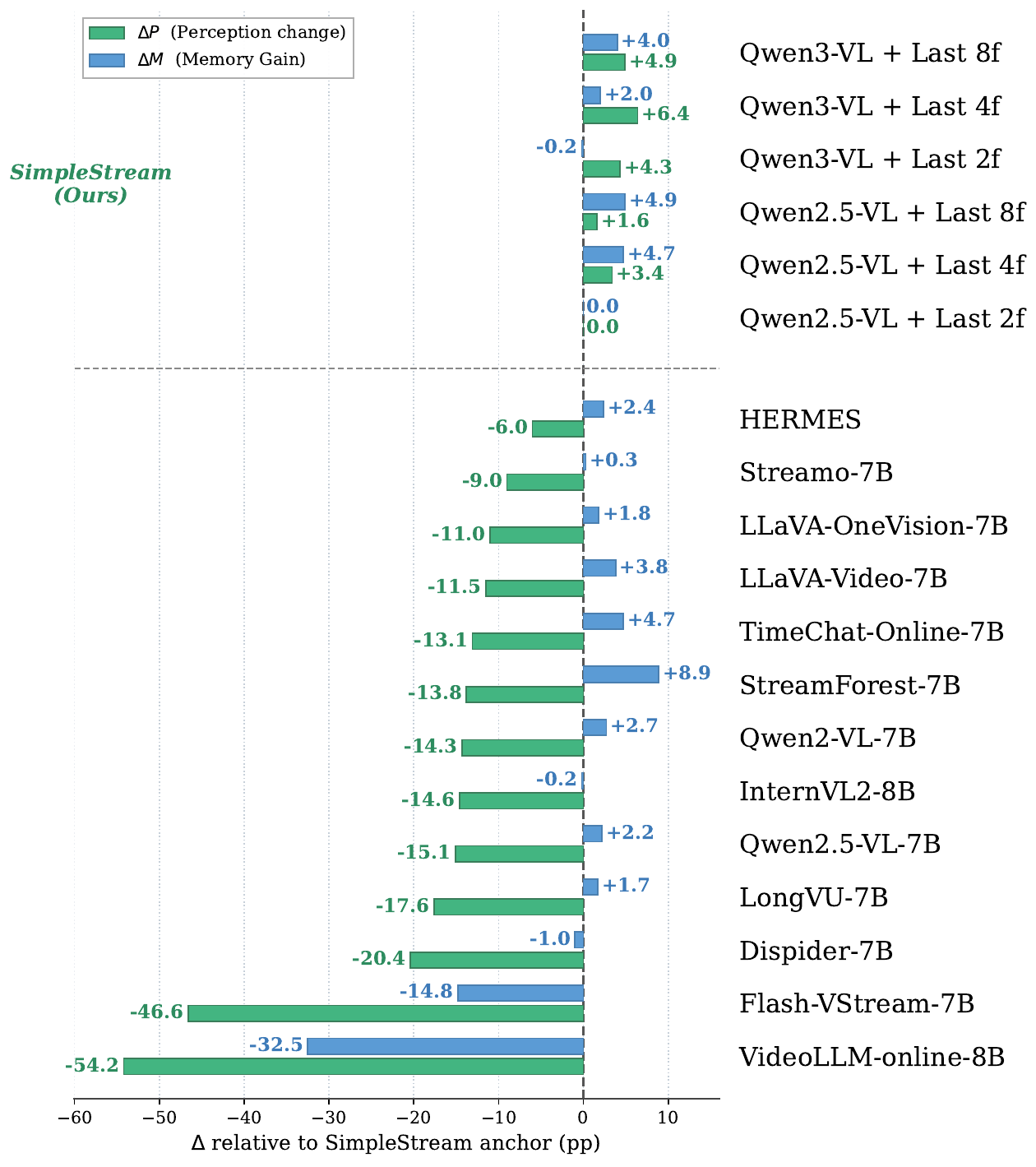}
  \caption{\textbf{Perception cost ($\Delta P$) and Memory Gain
  ($\Delta M$) relative to the \textsc{SimpleStream} Qwen2.5-VL + 2f
  configuration.} Green bars show changes in OVO-Bench Real-Time average. Blue bars
  show changes on the mean of EPM and ASI. Relative to this short-window
  reference, many methods improve memory, but external baselines still
  incur a substantial perception cost.}
  \label{fig:perception_fidelity}
  \vspace{-1.0\baselineskip}
\end{wrapfigure}

For any streaming method, we measure the change in real-time perception as
\begin{equation}
\Delta P = \mathrm{RT}_{\text{method}} - \mathrm{RT}_{\textsc{SimpleStream}}
\label{eq:delta_p}
\end{equation}
where $\mathrm{RT}$ is the OVO-Bench Real-Time category average. To quantify the
memory side, we define \emph{Memory Gain} as the change in the mean of EPM and
ASI, the two Backward tracks most directly tied to recalling previously
observed events:
\begin{equation}
\Delta M = \mathrm{ER}_{\text{method}} - \mathrm{ER}_{\textsc{SimpleStream}}
\label{eq:delta_m}
\end{equation}
where $\mathrm{ER} = (\mathrm{EPM} + \mathrm{ASI})/2$.
These two quantities make the trade-off explicit: $\Delta M > 0$ indicates
memory gain, while $\Delta P < 0$ indicates a perception cost. For
descriptive cross-method visualization, Figure~\ref{fig:perception_fidelity}
uses the \textsc{SimpleStream} Qwen2.5-VL + 2f configuration as a common
reference under the unified protocol.

Figure~\ref{fig:perception_fidelity} shows that the dominant pattern is still
perception loss: every evaluated external baseline falls below \textsc{SimpleStream}
on $\Delta P$. This holds across memory-bank methods, retrieval-based
augmentation, and offline long-context baselines, suggesting that the
perception cost is broad rather than architecture-specific. In contrast,
positive memory gain becomes common once memory is measured on EPM and ASI.
Among published streaming systems, StreamForest shows the clearest
memory-side gain ($\Delta M = +8.9$), but it pays a much larger perception
penalty ($\Delta P = -13.8$). HERMES also gains on memory
($\Delta M = +2.4$), yet still incurs a substantial perception cost
($\Delta P = -6.0$).

Our controlled Visual-RAG ablation shows the same asymmetry even more directly.
Retrieving historical chunks improves EPM and ASI by 6.6 points on average,
but reduces real-time perception by 4.9 points relative to the matched recent
window. Taken together, these results support a perception-memory trade-off
under a cleaner definition of memory: current mechanisms can improve
memory-oriented behavior, but these gains are typically purchased with a
broader and more consistent loss in present-scene perception. The central
challenge is therefore not simply to preserve more history, but to recover
useful past evidence without corrupting the model's perception of the present.

\subsection{Benchmark Limitations}
\paragraph{HLD is not a memory task.}
We argue that HLD is conceptually misaligned with long-term memory evaluation.
Hallucination detection primarily measures whether the model resists misleading
prompts, unsupported claims, or semantically inconsistent options. This ability
is related to robustness and grounded verification, but it is not equivalent to
recalling previously observed events from a long video stream. A model can fail
HLD without forgetting the past, just as it can succeed on HLD without
demonstrating strong episodic memory. Placing HLD under Backward Tracing
therefore conflates two distinct abilities: memory recall and hallucination
robustness. In our Visual-RAG study, for example, HLD drops by 12.4 points even
when memory-oriented tracks such as EPM and ASI improve. We therefore caution
against interpreting HLD as a direct measure of long-range memory.

\paragraph{Macro-average favors perception-heavy gains.}
OVO-Bench reports a macro-average over 12 tracks, but these tracks are not
balanced across capability types. Real-Time Visual Perception occupies 6 tracks,
Backward Tracing occupies 3, and Forward Active Responding occupies 3. As a
result, equal weighting at the track level does not produce balanced weighting
at the capability level. Aggregate scores are therefore more sensitive to
changes on perception-oriented tracks than to comparable changes on
memory-oriented ones. This matters when evaluating streaming methods that trade
stronger access to history for weaker present-scene perception, because even
modest degradation on Real-Time tracks can dominate the final score. We
therefore interpret the macro-average together with per-track results rather
than treating it as a neutral summary of overall streaming ability.

\FloatBarrier

\section{Why Does a Simple Baseline Win?}
\label{sec:discussion_rec}

\paragraph{Recent context matters most. }
Our results suggest that the main strength of current streaming VLMs does not
come from increasingly elaborate memory mechanisms, but from their already
strong short-horizon perception. Modern VLM backbones can read text, recognize
objects, track local actions, and answer query-conditioned questions well when
the recent visual evidence remains clear. In this regime, preserving an
undiluted view of the latest frames is often more valuable than injecting
additional historical summaries whose relevance is uncertain. Model scale does
affect how much recent context a backbone can use effectively: stronger models
can sometimes benefit from somewhat wider recent windows. But this is still a
recent-context effect rather than evidence that complex memory is necessary.
This perspective helps explain why \textsc{SimpleStream} is competitive: it
protects the signal that current backbones use best, namely dense and reliable
recent visual evidence.

\paragraph{Complex memory can hurt present perception.}
The converse is equally important: memory is not free. Compression, retrieval
noise, abstract latent states, or large memory injection can all interfere with
the model's understanding of the current scene, even when they are intended to
help long-range reasoning. In other words, adding more history can reduce the
effective clarity of the present input. One plausible mechanism is attention
dilution: when too much retrieved or summarized context is injected, the model
may allocate capacity away from the most relevant recent evidence. We do not
claim this mechanism as an established empirical fact here, but it offers a
useful hypothesis for why stronger memory modules can still degrade real-time
perception.
This does not argue against memory-centric designs in principle;
rather, it argues for evaluation that reveals when they improve
recall-oriented behavior without imposing unacceptable costs on
present-scene perception.

\paragraph{Benchmark design further amplifies this advantage.}
This advantage is also partly structural. Current benchmarks do not purely
measure long-term memory; instead, their aggregate scores still place heavy
weight on recent perception. As a result, methods that preserve clear recent visual evidence can gain twice: they align with the strongest capability of today's
backbones, and they are rewarded by evaluation protocols that overweight that
capability in the final score. From this perspective, \textsc{SimpleStream}
wins not only because it is strong, but also because the benchmark favors the
kind of strength it preserves. This does not invalidate the result, but it does
change its interpretation: benchmark leadership is not the same thing as
solving long-horizon memory.

\paragraph{Implications for future research.}
These observations suggest two concrete directions. For future \emph{models}, a
promising principle is \emph{recent-first, history-on-demand}: preserve the
recent context by default, and access historical memory only when the current
evidence is insufficient. More broadly, memory modules should be evaluated not
only by recall gains, but also by whether they damage real-time perception. For
future \emph{benchmarks}, evaluation should explicitly separate perception,
memory recall, and hallucination robustness rather than collapsing them into a
single macro-average. More transparent reporting of the
accuracy-efficiency trade-off would make it easier to distinguish genuinely
better long-context reasoning from methods that simply preserve the benchmark's
most rewarded capability.

\section{Conclusion}
\label{sec:conclusion}

We show that \textsc{SimpleStream}, a simple baseline, is already strong enough to exceed
recently published complex-memory streaming systems on both OVO-Bench and
StreamingBench while remaining latency-competitive.
Our results show that additional memory, retrieval, or compression
should be justified by clear gains over \textsc{SimpleStream} on the
relevant capability slices.
Our analyses show why this baseline is so competitive. Current memory
injection techniques often improve recall-oriented cases at the cost of
current-scene perception, producing a consistent perception-memory
trade-off that we quantify with real-time perception change ($\Delta P$).
Moreover, performance often saturates with a small recent window.
Although some higher-capacity backbones benefit from longer recent
context, the optimal window does not grow monotonically with model
scale, reinforcing that more history is not always better and that
current benchmarks may not faithfully reward long-term memory as
intended.
We therefore recommend that future work report strong simple baselines,
disaggregated perception-versus-memory metrics, and transparent
efficiency statistics before claiming progress.
The central open problem is not how to add more memory, but how to use
history without degrading current-scene understanding.

\section{Limitations}
\label{sec:limitations}

\paragraph{Dependence on strong backbone families.}
\textsc{SimpleStream} is evaluated on top of strong modern VLM backbones,
specifically Qwen2.5-VL and Qwen3-VL. As a result, our conclusions are coupled
to the capabilities of this backbone family: strong recent-context performance
may partly reflect the fact that these models already provide robust short-range
perception, OCR, and query-conditioned reasoning. We therefore do not claim
that the same degree of competitiveness will automatically transfer to broader
model families with different pretraining data, visual encoders, or temporal
reasoning characteristics. Extending the comparison to a wider range of
backbones is an important direction for future work.
\paragraph{Scope as a strong-baseline paper.}
This paper is deliberately positioned as a \emph{strong baseline} study rather
than a proposal of a new streaming video understanding architecture. Our main
contributions are to establish \textsc{SimpleStream} as a strong baseline,
clarify how results on memory-heavy benchmarks should be interpreted, and analyze the
accuracy-efficiency and perception-memory trade-offs. Accordingly,
\textsc{SimpleStream} does not introduce a new memory-centric architecture, a
new long-term memory mechanism, or a new retrieval/compression design. The
paper therefore should not be read as solving long-horizon video understanding;
rather, it clarifies what current benchmarks already reward and what future
memory-centric methods must surpass under stronger controls.

\bibliography{main}
\bibliographystyle{bibstyle}

\appendix


\end{document}